\documentclass[lettersize,journal]{IEEEtran}
\usepackage{amsmath,amsfonts}
\usepackage{algorithmic}
\usepackage{algorithm}
\usepackage{array}
\usepackage[caption=false,font=normalsize,labelfont=sf,textfont=sf]{subfig}
\usepackage{textcomp}
\usepackage{stfloats}
\usepackage{url}
\usepackage{verbatim}
\usepackage{graphicx}
\usepackage{cite}
\usepackage{scalerel}
\usepackage{diagbox}
\usepackage{booktabs,multirow}
\usepackage{xcolor}
\usepackage{nicematrix}
\usepackage{tikz}
\definecolor{orcidlogocol}{HTML}{A6CE39}
\tikzset{
  orcidlogo/.pic={
    \fill[orcidlogocol] svg{M256,128c0,70.7-57.3,128-128,128C57.3,256,0,198.7,0,128C0,57.3,57.3,0,128,0C198.7,0,256,57.3,256,128z};
    \fill[white] svg{M86.3,186.2H70.9V79.1h15.4v48.4V186.2z}                svg{M108.9,79.1h41.6c39.6,0,57,28.3,57,53.6c0,27.5-21.5,53.6-56.8,53.6h-41.8V79.1z M124.3,172.4h24.5c34.9,0,42.9-26.5,42.9-39.7c0-21.5-13.7-39.7-43.7-39.7h-23.7V172.4z}
                 svg{M88.7,56.8c0,5.5-4.5,10.1-10.1,10.1c-5.6,0-10.1-4.6-10.1-10.1c0-5.6,4.5-10.1,10.1-10.1C84.2,46.7,88.7,51.3,88.7,56.8z};
  }
}
\newcommand\orcidicon[1]{\href{https://orcid.org/#1}{\mbox{\scalerel*{
\begin{tikzpicture}[yscale=-1,transform shape]
\pic{orcidlogo};
\end{tikzpicture}
}{|}}}}
\usepackage{tikz}
\usetikzlibrary{svg.path}
\usepackage{hyperref}
\usepackage{multirow}
\usepackage{adjustbox}
\usepackage{array}
\usepackage[section]{placeins}

\begin{document}

\title{Towards Accurate State Estimation: Motion Dynamics Kalman Filter for 3D Multi-Object Tracking}

\author{
Mohamed Nagy%
\IEEEauthorrefmark{1}\IEEEauthorrefmark{2}%
\orcidicon{0000-0002-5309-064X},~
\IEEEmembership{Member,~IEEE,}
Naoufel Werghi%
\IEEEauthorrefmark{1}%
\orcidicon{0000-0002-5542-448X},~
\IEEEmembership{Senior Member,~IEEE,}
Bilal Hassan\IEEEauthorrefmark{3}\orcidicon{0000-0003-3672-8100},~\IEEEmembership{Member,~IEEE,}
Jorge Dias%
\IEEEauthorrefmark{1}%
\orcidicon{0000-0002-2725-8867},~
\IEEEmembership{Senior Member,~IEEE,}
and Majid Khonji%
\IEEEauthorrefmark{1}%
\orcidicon{0000-0002-1548-9377},~
\IEEEmembership{Member,~IEEE}

\thanks{This research was supported by the Center for Autonomous Robotic Systems (KU-CARS), Khalifa University of Science and Technology.}

\thanks{\IEEEauthorrefmark{1}M. Nagy, N. Werghi, J. Dias, and
M. Khonji are with the Department of Computer Science,
Khalifa University, Abu Dhabi, United Arab Emirates
(e-mail: mohamed.nagy@ieee.org; naoufel.werghi@ku.ac.ae;
jorge.dias@ku.ac.ae; majid.khonji@ku.ac.ae).}

\thanks{\IEEEauthorrefmark{2}M. Nagy was with the Technical University of Munich (TUM), School of Engineering and Design, Department of Mobility Systems Engineering, Institute of Automotive Technology, and Munich Institute of Robotics and Machine Intelligence (MIRMI), 85748 Garching, Germany, from January to June 2026, where the autonomous-vehicle experiments reported in this article were conducted}

\thanks{\IEEEauthorrefmark{3}Author is with the Division of Engineering, New York University Abu Dhabi, 129188, Abu Dhabi, UAE (e-mail: 
bilal.hassan@nyu.edu).}

}

\markboth{This work has been submitted to the IEEE for possible publication. Copyright may be transferred without notice.}%
{Shell \MakeLowercase{\textit{et al.}}: A Sample Article Using IEEEtran.cls for IEEE Journals}


\maketitle

\begin{abstract}
Robust 3D state estimation in multi-object tracking is crucial for object localization in self-driving vehicles. Misalignment between the Kalman filter motion model and an object's motion can cause state drift, particularly in distant and occlusion cases. The static motion model assumption in the conventional Kalman filter is widely used in 3D multi-object tracking due to its low latency; however, its estimation accuracy degrades as the object's motion disaligns with the chosen model due to traffic. Although recent work introduces a multimodel Kalman filter that runs multiple motion models simultaneously to improve state robustness, it incurs significant computational overhead, so many studies still favor the conventional Kalman filter to maintain efficiency. This work addresses this trade-off by introducing the motion dynamics Kalman filter that accounts for motion variation through temporal modeling of object motion and adaptively scales the contribution of the motion components in the motion model to correct state-prediction drift caused by this variation, while maintaining computational efficiency by retaining a single-model structure. Evaluations on KITTI and the Waymo Open Dataset, complemented by vehicle experiments, show that the proposed filter achieves tracking performance comparable to the multimodel Kalman filter with an average 1.72x speedup, while outperforming it in handling highly distant and prolonged occluded objects.
\end{abstract}


\section{Introduction}
State estimation in multi-object tracking (MOT) is vital for autonomous driving applications for motion modeling and trajectory prediction to prevent potential collisions. Given an object's current location (measurement), motion-based MOT methods employ Bayesian filters, such as the Kalman Filter~\cite{10.1115/1.3662552} (KF), to estimate its subsequent position (state) with uncertainty, a process known as \emph{state estimation}. When a new measurement comes, the predicted state is refined through a \emph{state update} stage. The conventional KF (Baseline KF) relies on a static (constant) motion model, such as constant acceleration, for next-state prediction, which is widely used in recent work~\cite{nagy2024robmotrobust3dmultiobject,nagy,CasTrack,Rethink_mot} due to its low latency. 

\begin{figure}[!t]
    \centering
    \includegraphics[width=\linewidth]{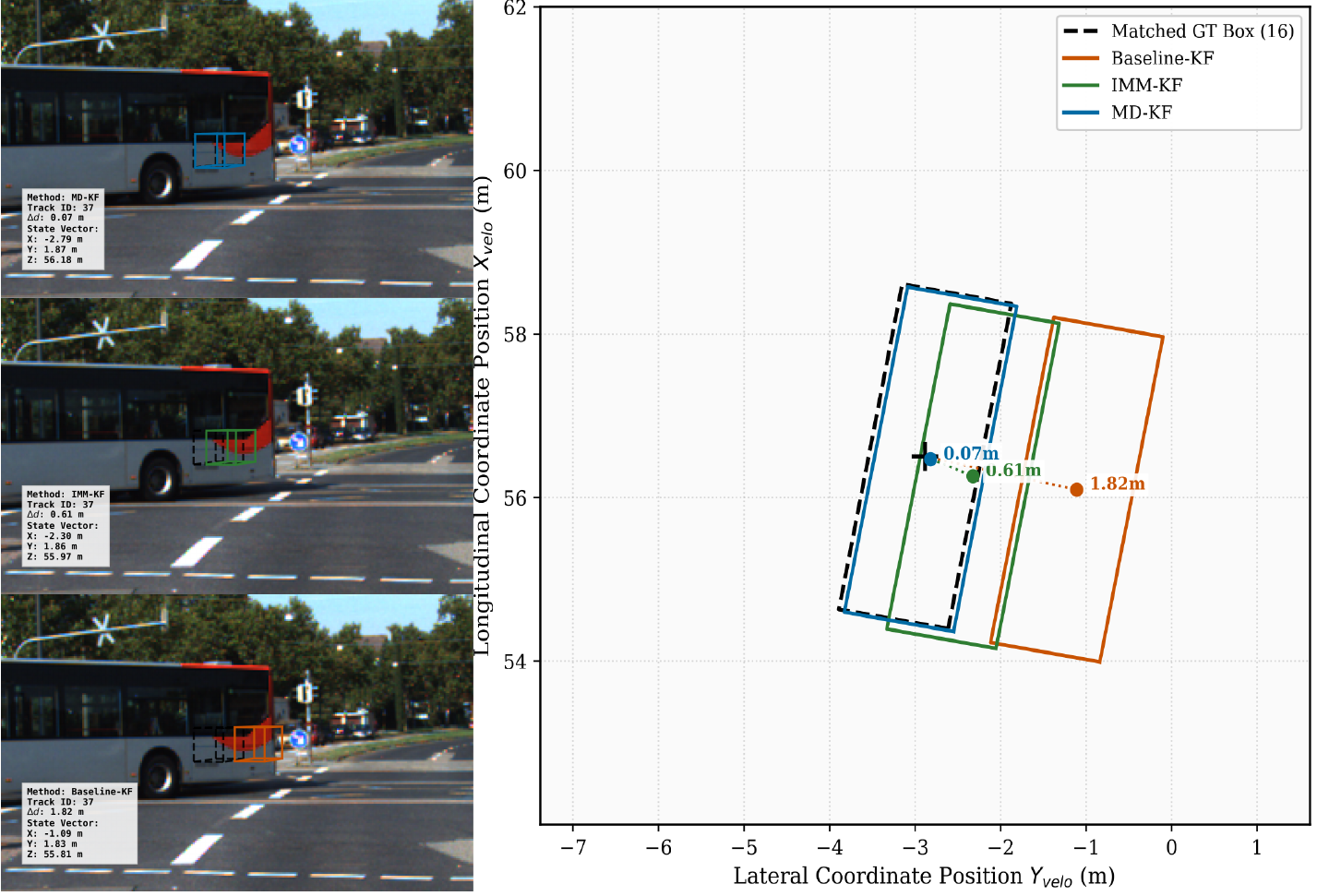}
    \caption{A car is stopped at a traffic light, and a bus blocks it in the LiDAR point cloud. The baseline KF's state prediction (orange bounding box) drifts far from the car's location, while the IMM-KF (green bounding box) drifts less than the baseline KF. The MD-KF state prediction (blue bounding box) aligns with the ground-truth (disconnected-line bounding box)  car location, showing greater robustness to object occlusion than the other models. }
    \label{fig:occlusion_with_bus_base}
\end{figure}

However, a static motion model can inject noise into the predicted state if the selected model does not match the object's current motion, particularly in high-traffic environments. Figure~\ref{fig:occlusion_with_bus_base} shows a state prediction of an occluded stationary car using a baseline KF with a constant-jerk motion model, which drifts away from the car's location. Recent work~\cite{imm_kf_mot24} introduces an interactive multimodel Kalman filter (IMM-KF) in MOT that runs multiple motion models simultaneously to account for motion misalignment. However, it comes at the cost of KF efficiency because it handles multiple models. Yet, IMM-KF may still drift significantly with stationary (Figure~\ref{fig:occlusion_with_bus_base})  and noisy detections (Figure~\ref{fig:bev-realtime-tracking}). 

To address this trade-off, we introduce the motion dynamics Kalman filter (MD-KF) that incorporates variation in an object’s motion and adaptively corrects its state prediction. MD-KF captures variation in object motion across consecutive measurements to quantify variation in each motion-state component, then uses it to correct the state prediction to account for motion changes and mitigate the state drift caused by that variation. This formulation retains the efficiency of the baseline KF while achieving competitive robustness in state estimation with IMM-KF, as shown in Figure~\ref{fig:occlusion_with_bus_base}. 

The contributions of this work are summarized as follows.

\begin{itemize}
    \item Proposing MD-KF, which quantifies motion variation in objects across consecutive frames and the adaptive correctness of the predicted state while preserving the motion model’s singularity.  

 \item Demonstrating comparative state-estimation performance against IMM-KF~\cite{imm_kf_mot24} across multiple detectors and datasets, particularly under challenging tracking conditions such as occlusions and distant objects with interrupted detections.

\item By retaining a single adaptive motion model, MD-KF performs state updates and state predictions at an average throughput $1.72\times$ that of IMM-KF.

\item A vehicular experiment on the EDGAR~\footnote{https://www.mos.ed.tum.de/ftm/forschungsfelder/team-av-perception/edgar/}~\cite{karle2024edgarautonomousdrivingresearch} stack under rainy weather conditions shows the MD-KF's state-estimation robustness and efficiency.
\end{itemize}
\section{Related Work}
% General paragraph about motion models and localization state estimation

The literature presents different findings regarding the most suitable motion model for state prediction. Constant-velocity models remain widely adopted because of their simplicity and low
computational cost~\cite{wang2021immortal,Pang2021SimpleTrackUA,he20243d,zhu2022msa,Rethink_mot,intro_tra_by_det_kim_eager,Weng20193DMT,scheidegger2018mono}. Some studies~\cite{yan2013maneuvering,na2022adaptive} further report that constant velocity can outperform higher-order models when target acceleration is small, or measurements are noisy. Conversely, acceleration-based models have improved prediction for maneuvering objects and during extended occlusions, where errors from the constant-velocity assumption may accumulate~\cite{nagy,CasTrack,9626850}. These findings indicate that a fixed motion model's suitability depends strongly on the target's observed motion.

Recent work has therefore investigated adaptive state prediction. Liu \emph{et al.}~\cite{11249796} proposed an adaptive estimation smoothing strategy that extrapolates the next acceleration from two consecutive acceleration estimates. The resulting position and velocity estimates are then smoothed with those from a conventional motion model. This approach accommodates variable acceleration. However, globally selected differential and smoothing factors govern its adaptation, which depends on the most recent acceleration difference. It does not independently quantify object- and direction-specific evidence for velocity, acceleration, and jerk over a temporal observation window.

Alternatively, recent methods~\cite{sani2024sensor,na2022adaptive, 11246301} employ an interacting multiple-model Kalman filter (IMM-KF), which operates multiple motion models in parallel and combines their state estimates according to recursively updated model probabilities. Although IMM-KF can adapt to different motion regimes, maintaining and mixing multiple state estimates and covariance matrices for every tracked object introduces considerable computational overhead in multi-object tracking.

To address these limitations, we propose MD-KF, which quantifies object- and direction-specific motion variation over a temporal window and uses it to adjust the predicted state adaptively. MD-KF thereby mitigates state deviation caused by motion changes while retaining the computational efficiency of a single-model KF.

\section{Method}
\begin{figure*}[!t]
    \centering
    \includegraphics[width=\linewidth]{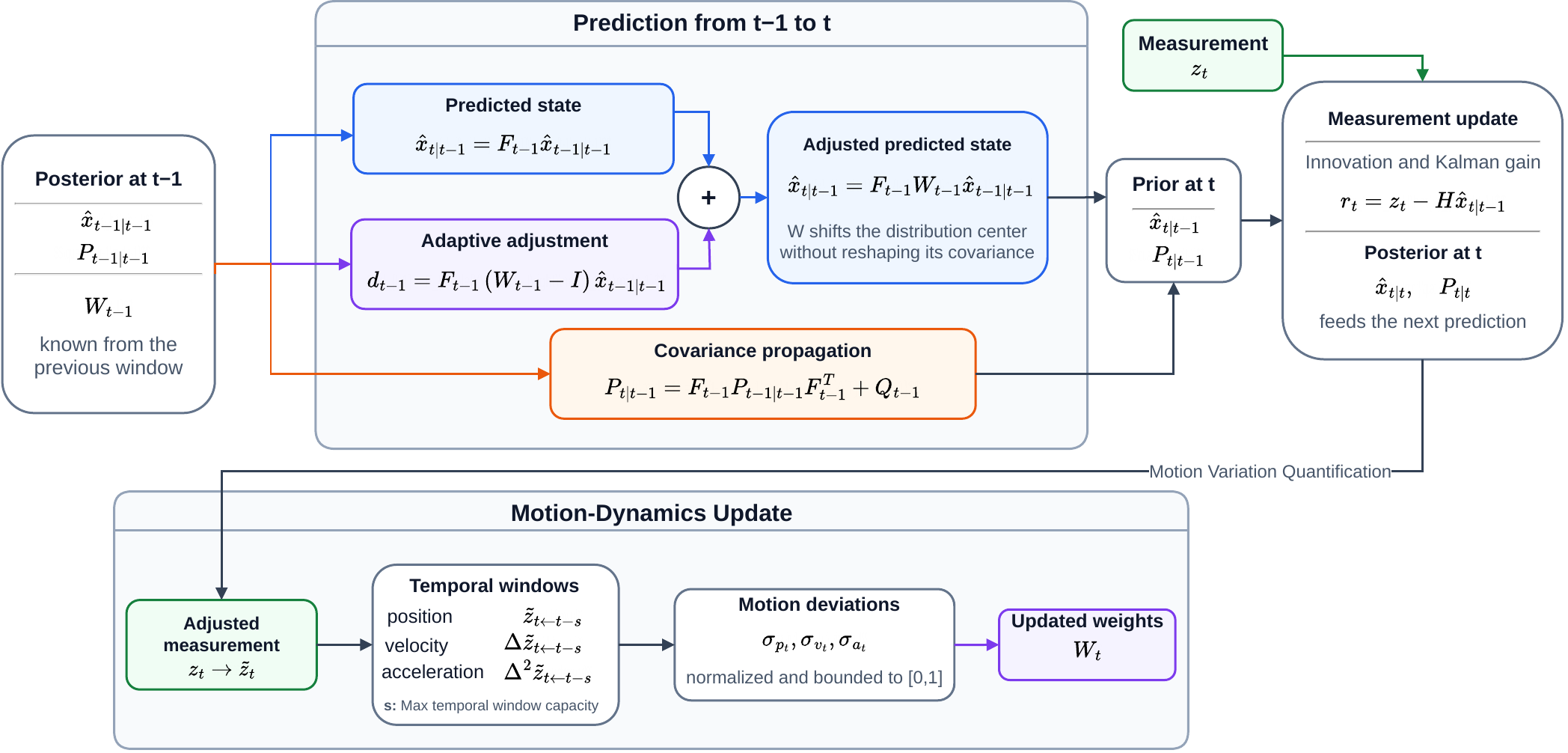}
    \caption{\label{fig:main_figure} An overview of the MD-KF pipeline that integrates motion dynamics into the baseline KF. As a measurement $z_t$ arrives, the adjusted measurement $\tilde z_t$ is incorporated into three temporal windows (position, velocity, and acceleration) to quantify the change in each component. The motion variation weights are then obtained from the standard deviation across the temporal windows and normalized using the MD-KF normalization factors to form the updated motion dynamics weights $W_t$.  The updated motion variation matrix $W_t$ is then used for next-state prediction by scaling the contribution of motion components in the baseline KF motion model through the adaptive adjustment step.}
\end{figure*}

Figure~\ref{fig:main_figure} shows a high-level pipeline for incorporating motion dynamics into the conventional KF. 

Given the posterior estimate of the object state $\hat{x}_{t-1\mid t-1}$ at time $t-1$ and its covariance $P_{t-1\mid t-1}$, the state estimate $\hat{x}_{t\mid t-1}$ at time $t$ is obtained from the conventional KF (Predicted state in Figure~\ref{fig:main_figure}). MD-KF uses the motion-dynamics weights $\mathbf{W}_{t-1}$, computed from the recent motion history available at time $t-1$, to construct an adaptive adjustment $\mathbf{d}_{t-1}$ to the conventional state prediction. The adjustment scales the contribution of each motion component according to the motion recently exhibited by the individual object (Adaptive adjustment in Figure~\ref{fig:main_figure}) to the conventional KF predicted state (Adjusted predicted state in Figure~\ref{fig:main_figure}). Section~\ref{subsec:md_in_state_prediction} details this prediction-state procedure.

When a new measurement $\mathbf{z}_t$ becomes available, the innovation $r_t$
between the measurement and the predicted observation is computed and used in the conventional KF measurement update
to obtain the posterior state estimate $\hat{\mathbf{x}}_{t|t}$ and its
covariance $\mathbf{P}_{t|t}$ (Measurement update and Posterior in Figure~\ref{fig:main_figure}).

The new measurement then updates the object's motion variation over a temporal window to obtain the updated variation-noise weights $W_{t}$ (Motion-Dynamics Update in Figure~\ref{fig:main_figure}), which are used to adjust motion variation in the next state prediction. Section~\ref{subsec:state_update} details this process.

\subsection{Motion Dynamics in State Prediction}
\label{subsec:md_in_state_prediction}

Let \(\hat{\mathbf{x}}_{t-1|t-1}\) and \(\mathbf{P}_{t-1|t-1}\) denote the posterior state estimate and its error covariance obtained from measurement $z_{t-1}$ at time $t-1$ . The state vector is defined as

\[
\mathbf{x}_{t}
=
[p_{x,t},p_{y,t},v_{x,t},v_{y,t},
a_{x,t},a_{y,t},j_{x,t},j_{y,t}]^\top.
\]

The transition matrix \(\mathbf{F}_t\) is used to propagate the position, velocity, acceleration, and jerk between consecutive frames over the time interval \(\Delta t\), which is used to estimate the next state  \(\hat{\mathbf{x}}_{t|t-1}\).

\[
\hat{\mathbf{x}}_{t|t-1}
=
\mathbf{F}_t\hat{\mathbf{x}}_{t-1|t-1}.
\]

MD-KF obtains the motion variation (motion dynamics) from the last state update at $t-1$, as described in Section~\ref{subsec:state_update}, in a diagonal matrix form. 

\begin{equation}
\scriptsize
    \mathbf{W}_{t-1}
=
\operatorname{diag}
\left(
1,1,
w^v_{x,t-1},w^v_{y,t-1},
w^a_{x,t-1},w^a_{y,t-1},
w^j_{x,t-1},w^j_{y,t-1}
\right)
\end{equation}

The coefficients $(w^v_{x,t-1},w^v_{y,t-1})$ regulate the velocity components, $(w^a_{x,t-1},w^a_{y,t-1})$ regulate the acceleration components, and $(w^j_{x,t-1},w^j_{y,t-1})$ regulate the jerk components. Each coefficient lies in \([0,1]\) and controls the contribution of the corresponding motion component to the next prediction. 

MD-KF incorporates the weighted motion dynamics $W_{t-1}$ as an adjustment to the predicted state rather than a transition. Let

\[
\mathcal{Z}_{t-1}
=
\{\mathbf{z}_1,\ldots,\mathbf{z}_{t-1}\}
\]

denote the observation history available through time \(t-1\). Since \(\mathbf{W}_{t-1}\) and \(\hat{\mathbf{x}}_{t-1|t-1}\) are both calculated from information available in \(\mathcal{Z}_{t-1}\), they are treated as known during the prediction of the state at time \(t\).

MD-KF introduces the adaptive prediction-adjustment vector

\begin{equation}
    \mathbf{d}_{t-1}
=
\mathbf{F}_t
\left(
\mathbf{W}_{t-1}-\mathbf{I}
\right)
\hat{\mathbf{x}}_{t-1|t-1}
\end{equation}

Equivalently, \(\mathbf{d}_{t-1}\) is the difference between the motion-dynamics-adjusted and the predicted state:

\[
\mathbf{d}_{t-1}
=
\mathbf{F}_{t}\mathbf{W}_{t-1}\hat{\mathbf{x}}_{t-1|t-1}
-
\mathbf{F}_{t}\hat{\mathbf{x}}_{t-1|t-1}.
\]
This expression provides the following intuition. When \(\mathbf{W}_{t-1}=\mathbf{I}\), the adjustment is zero. As a motion-dynamics coefficient approaches zero, its associated state component contributes less to the predicted state estimate. Thus, \(\mathbf{d}_{t-1}\) adapts the prediction according to the recent motion exhibited by the individual object without maintaining multiple parallel motion models.

Let $\mathbf{X}_{t} \in \mathbb{R}^{8}$ denote the random variable
representing the unknown true kinematic state of an object at time $t$. Then, the prediction model is written as 
\[
\mathbf{X}_{t}
=
\mathbf{F}_t\mathbf{X}_{t-1}
+
\mathbf{d}_{t-1}
+
\mathbf{q}_{t-1},
\]

The process noise is assumed to be conditionally zero-mean and
uncorrelated with the previous posterior estimation error:
\[
\mathbb{E}[
\mathbf{q}_{t-1}\mid\mathcal{Z}_{t-1}
]
=
\mathbf{0},
\qquad
\mathbb{E}[
\mathbf{q}_{t-1}\mathbf{q}_{t-1}^{\mathsf T}
]
=
\mathbf{Q}_{t-1}.
\]

Taking the conditional expectation given \(\mathcal{Z}_{t-1}\) gives

\[
\begin{aligned}
\hat{\mathbf{x}}_{t|t-1}
&=
\mathbb{E}
\left[
\mathbf{X}_{t-1}
\mid
\mathcal{Z}_{t-1}
\right]\\
&=
\mathbf{F}_t\hat{\mathbf{x}}_{t-1|t-1}
+
\mathbf{d}_{t-1}\\
&=
\mathbf{F}_t\hat{\mathbf{x}}_{t-1|t-1}
+
\mathbf{F}_t
\left(
\mathbf{W}_{t-1}-\mathbf{I}
\right)
\hat{\mathbf{x}}_{t-1|t-1}\\
&=
\mathbf{F}_t\mathbf{W}_{t-1}
\hat{\mathbf{x}}_{t-1|t-1}.
\end{aligned}
\]

Accordingly, the state prediction used in MD-KF is

\begin{equation}
    \hat{\mathbf{x}}_{t|t-1}
=
\mathbf{F}_t\mathbf{W}_{t-1}
\hat{\mathbf{x}}_{t-1|t-1}
\end{equation}

To derive the effect of the adaptive adjustment on the prediction
error and covariance propagation, define the prior estimation error as 

\[
\mathbf{e}_{t|t-1}
=
\mathbf{X}_{t}-\hat{\mathbf{x}}_{t|t-1}
\]

Substituting the conditional state model and predicted mean yields

\[
\begin{aligned}
\mathbf{e}_{t|t-1}
&=
\mathbf{F}_t\mathbf{X}_{t-1}
+
\mathbf{d}_{t-1}
+
\mathbf{q}_{t-1}
-
\left(
\mathbf{F}_t\hat{\mathbf{x}}_{t-1|t-1}
+
\mathbf{d}_{t-1}
\right)\\
&=
\mathbf{F}_t
\left(
\mathbf{X}_{t-1}-\hat{\mathbf{x}}_{t-1|t-1}
\right)
+
\mathbf{q}_{t-1}\\
&=
\mathbf{F}_t\mathbf{e}_{t-1|t-1}
+
\mathbf{q}_{t-1}.
\end{aligned}
\]

The deterministic adjustment \(\mathbf{d}_{t-1}\) cancels from the error equation; therefore, \(\mathbf{W}_{t-1}\) changes the center of the predicted distribution but does not directly transform the estimation error.

Given the predicted state covariance is
\[
\mathbf{P}_{t|t-1}
=
\mathbb{E}
\left[
\mathbf{e}_{t|t-1}
\mathbf{e}_{t|t-1}^{\top}
\mid
\mathcal{Z}_{t-1}
\right].
\]
The proposed formulation therefore retains the conventional KF covariance propagation while adapting the predicted state estimate to the object-specific motion observed in the recent temporal window.

\subsection{Motion-Dynamics Weight Update}
\label{subsec:state_update}
To estimate the object motion dynamics, a localization term that closely
represents the object position is required. We consider the
detector-specific localization-noise covariance $\mathbf{D}_t$ estimated
in~\cite{nagy2024robmotrobust3dmultiobject}. Building on this noise
model, we define a post-measurement motion observation as
\begin{equation}
\tilde{\mathbf{z}}_t
=
\mathbf{z}_t
-
(\mathbf{H}\mathbf{K}_t)
\operatorname{diag}(\mathbf{D}_t),
\label{eq:post_measurement_observation}
\end{equation}
where $\mathbf{H}\mathbf{K}_t$ maps the uncertainty-dependent correction
to the measurement space. The resulting
$\tilde{\mathbf{z}}_t$ is used only for estimating the motion-dynamics
weights and does not replace the measurement used in the conventional
KF update.

Since the measurement provides the current object position, its temporal
changes are represented using first- and second-order finite
differences:
\begin{equation}
\begin{aligned}
\text{First-order difference:}\quad
\Delta\tilde{\mathbf{z}}_t
&=
\tilde{\mathbf{z}}_t-\tilde{\mathbf{z}}_{t-1},
\\
\text{Second-order difference:}\quad
\Delta^2\tilde{\mathbf{z}}_t
&=
\Delta\tilde{\mathbf{z}}_t
-
\Delta\tilde{\mathbf{z}}_{t-1}.
\end{aligned}
\label{eq:dynamic_change_delta}
\end{equation}
For a fixed sampling interval, the first- and second-order position
differences provide velocity-related and acceleration-related motion
evidence, respectively.

MD-KF quantifies the variation of motion components across consecutive measurements. It requires at least two measurements to quantify a motion component's variation and uses a tunable maximum window length $l$. Define the position, velocity, and acceleration windows as
\begin{equation}
\begin{aligned}
    \mathcal P_t
&=
\left\{
\tilde{\mathbf z}_{t-l_p},
\ldots,
\tilde{\mathbf z}_t
\right\}\\
    \mathcal V_t
&=
\left\{
\Delta \tilde{\mathbf z}_{t-l_v},
\ldots,
\Delta \tilde{\mathbf z}_t
\right\}\\
    \mathcal A_t
&=
\left\{
\Delta^2 \tilde{\mathbf  z}_{t-l_a},
\ldots,
\Delta^2 \tilde{\mathbf z}_t
\right\}
\end{aligned}
\end{equation}
, respectively. Since a higher-order motion component characterizes the temporal
variation of its preceding component, MD-KF uses the standard deviation
of each motion-component window as activation evidence for the next
higher-order component. Accordingly, $\sigma^p_{d,t}$ provides evidence
of velocity, $\sigma^v_{d,t}$ provides evidence of acceleration, and
$\sigma^a_{d,t}$ provides evidence of jerk along axis
$d\in\{x,y\}$.
\[ \sigma^p_{d,t} = \operatorname{Std}(\mathcal P_{d,t}), \]\[ \sigma^v_{d,t} = \operatorname{Std}(\mathcal V_{d,t}), \] \[ \sigma^a_{d,t} = \operatorname{Std}(\mathcal A_{d,t}). \]
For example, when a vehicle moves with approximately constant velocity, its frame-to-frame displacement $\Delta\tilde{\mathbf{z}}_t =\tilde{\mathbf{z}}_t-\tilde{\mathbf{z}}_{t-1}$ remains nearly constant. Consequently, the standard deviation of the corresponding displacement window satisfies $\sigma^v_{d,t}\approx 0$, indicating little temporal variation in velocity. Conversely, when the vehicle accelerates or decelerates, its frame-to-frame displacement changes over time, thereby increasing $\sigma^v_{d,t}$. This increased deviation indicates an acceleration component along axis $d\in\{x,y\}$.

The motion-dynamics evidence values
$\{\sigma^p_{d,t},\sigma^v_{d,t},\sigma^a_{d,t}\}$
are normalized and bounded to $[0,1]$ using tunable
motion-dynamics factors
$\{\gamma^p_d,\gamma^v_d,\gamma^a_d\}$.
Accordingly, the velocity, acceleration, and jerk weights are computed as
\[
w^v_{d,t}
=
\min\left(
1,\frac{\sigma^p_{d,t}}{\gamma^p}
\right),
\]
\[
w^a_{d,t}
=
\min\left(
1,\frac{\sigma^v_{d,t}}{\gamma^v}
\right),
\]
\[
w^j_{d,t}
=
\min\left(
1,\frac{\sigma^a_{d,t}}{\gamma^a}
\right).
\]

The evidence values and their corresponding normalization factors are collected into the vectors
\begin{equation}
\begin{aligned}
    \boldsymbol{\eta}_t
=&
\begin{bmatrix}
\sigma^p_{x,t} &
\sigma^p_{y,t} &
\sigma^v_{x,t} &
\sigma^v_{y,t} &
\sigma^a_{x,t} &
\sigma^a_{y,t}
\end{bmatrix}^{\mathsf T}\\
\boldsymbol{\gamma}
=&
\begin{bmatrix}
\gamma^p_x &
\gamma^p_y &
\gamma^v_x &
\gamma^v_y &
\gamma^a_x &
\gamma^a_y
\end{bmatrix}^{\mathsf T}.
\end{aligned}
\end{equation}

Then, the compact vector normalization can be written as
\begin{equation}
  \begin{aligned}
    \boldsymbol{\rho}_t
    &=
    \boldsymbol{\eta}_t
    \oslash\ 
    \boldsymbol{\gamma},
    \\
\boldsymbol{\omega}_t
&=
\frac{1}{2}
\left(
\mathbf 1_6
+
\boldsymbol{\rho}_t
-
\left|
\mathbf 1_6-\boldsymbol{\rho}_t
\right|
\right)
  \end{aligned}
  \label{eq:ch4:weight_update}
\end{equation}

where \(\oslash\) denotes element-wise division. The motion dynamics weight $W_t$ is then used for the next-state prediction.
\[
\mathbf W_t
=
\operatorname{diag}
\left(
1,1,\boldsymbol{\omega}_t^{\mathsf T}
\right).
\]

\section{Results}
\subsection{Experiment Protocol}
\label{sub:exp_protocol}
MD-KF is implemented in C++17 using the Eigen3 library and validated on a notebook machine with an Intel Core i9 processor and 48GB RAM; no GPU is involved. The tuning process of MD-KF is done on the KITTI~\cite{Geiger2012CVPR} training dataset using the CasA~\cite{casA_detector} detector, and the settings remain unchanged across all experiments. Table~\ref{tab:tuned_params} shows the tuned parameters. 

This work is evaluated by comparing the motion-based tracking performance of the baseline KF, IMM-KF~\cite{imm_kf_mot24}, and the proposed MD-KF on the unified benchmark tracking framework RobMOT~\cite{nagy2024robmotrobust3dmultiobject}. The method's tracking settings remain unified to ensure the performance gain comes directly from the included KF method. 

The baseline KF uses the constant-jerk transition model with the state vector defined in Section \ref{subsec:md_in_state_prediction}. The IMM-KF follows~\cite{imm_kf_mot24} and operates three parallel motion models: constant velocity, constant acceleration, and constant jerk. All three methods use the same measurements, measurement-noise settings, data-association procedure, and track-management configuration.

\begin{table}[!tbh]
    \centering
    \caption{\label{tab:tuned_params}Values of the tuned MD-KF parameters on the KITTI training dataset using the CasA detector. The parameters remain unchanged across all experiments conducted in this work.}
    \begin{tabular}{|c|c||c|c|}
    \hline
         Parameter& Value & Parameter & Value\\  
         \hline
         $\gamma^p_x$ & 0.14 & $\gamma^p_y$ & 0.14\\
         $\gamma^v_x$ & 0.60 & $\gamma^v_y$ & 0.60\\
         $\gamma^a_x$ & 0.28 & $\gamma^a_y$ & 0.28\\
          $l^p_x$ & 22 & $l^p_y$ & 22\\
         $l^v_x$ & 2 & $l^v_y$ & 2\\
         $l^a_x$ & 22 & $l^a_y$ & 22\\
         \hline
    \end{tabular}
\end{table}

\subsection{Performance Comparison on Benchmarks Dataset}
\label{sec:comp_dataset_benchmarks}
\subsubsection{Cross Detection Generalization Evaluation}
\begin{table*}[t]
\centering
\caption{Evaluation of the baseline KF, IMM-KF, and the proposed MD-KF on the KITTI validation dataset across five detectors. $\Delta$ shows the performance difference between the MD-KF and the corresponding method. The highest performance for each detector across the metrics is highlighted in \textbf{bold}. \textcolor{blue}{Blue} highlights MD-KF performance gains, while \textcolor{red}{red} highlights performance degradation. }
\label{tab:kitti-five-det-eval}
\small
\setlength{\tabcolsep}{5pt}
\begin{tabular}{|l|l|cccccc|}
\hline
Detector & Method & HOTA $\uparrow$& DetA $\uparrow$& AssA $\uparrow$& LocA $\uparrow$& MOTA $\uparrow$  & IDSW $\downarrow$ \\
\hline
\multirow{5}{*}{VirConv~\cite{VirConvDet_CVPR_23}}
 & Baseline KF     & 85.97 & 83.99 & 88.07 & 92.15 & 91.12 & \textbf{1} \\
 & IMM-KF & 85.95 & 83.96 & 88.08 & 92.13 & 91.07 & \textbf{1} \\
 & MD-KF  & \textbf{86.30} & \textbf{84.37} & \textbf{88.36}
          & \textbf{92.16} & \textbf{91.53} & \textbf{1} \\
          \hline
 & $\Delta$ MD-KF/Baseline KF
          & \textcolor{blue}{+0.33} & \textcolor{blue}{+0.38} & \textcolor{blue}{+0.29} & \textcolor{blue}{+0.01} & \textcolor{blue}{+0.41} & 0 \\
 & $\Delta$ MD-KF/IMM-KF
          & \textcolor{blue}{+0.35} & \textcolor{blue}{+0.41} & \textcolor{blue}{+0.28} & \textcolor{blue}{+0.03} & \textcolor{blue}{+0.46} & 0 \\
\hline
\multirow{5}{*}{CasA~\cite{casA_detector}}
 & Baseline KF     & 83.07 & 80.63 & 85.72 & 90.45 & 89.74 & \textbf{1} \\
 & IMM-KF & 82.99 & 80.57 & 85.62 & 90.45 & 89.59 & \textbf{1} \\
 & MD-KF  & \textbf{83.33} & \textbf{80.94} & \textbf{85.94}
          & \textbf{90.50} & \textbf{89.98} & \textbf{1} \\
          \hline
 & $\Delta$ MD-KF/Baseline KF
          & \textcolor{blue}{+0.26} & \textcolor{blue}{+0.31} & \textcolor{blue}{+0.22} & \textcolor{blue}{+0.05} & \textcolor{blue}{+0.24} & 0 \\
 & $\Delta$ MD-KF/IMM-KF
          & \textcolor{blue}{+0.34} & \textcolor{blue}{+0.37} & \textcolor{blue}{+0.32} & \textcolor{blue}{+0.05} & \textcolor{blue}{+0.39} & 0 \\
\hline
\multirow{5}{*}{PointRCNN~\cite{point_rcnn_detector}}
 & Baseline KF     & 77.42 & 73.30 & 82.01 & 87.31 & 84.81 & 6 \\
 & IMM-KF & 77.47 & 73.21 & 82.21 & \textbf{87.41} & 84.51 & \textbf{4} \\
 & MD-KF  & \textbf{77.71} & \textbf{73.39} & \textbf{82.52}
          & 87.40 & \textbf{84.86} & 5 \\
          \hline
 & $\Delta$ MD-KF/Baseline KF
          & \textcolor{blue}{+0.29} & \textcolor{blue}{+0.09} & \textcolor{blue}{+0.51} & \textcolor{blue}{+0.09} & \textcolor{blue}{+0.05} & \textcolor{blue}{$-1$} \\
 & $\Delta$ MD-KF/IMM-KF
          & \textcolor{blue}{+0.24} & \textcolor{blue}{+0.18} & \textcolor{blue}{+0.31} & \textcolor{red}{$-0.01$} & \textcolor{blue}{+0.35} & \textcolor{red}{+1} \\
\hline
\multirow{5}{*}{SECOND~\cite{second_detector}}
 & Baseline KF     & 78.42 & 74.76 & 82.53 & 88.44 & 85.56 & 3 \\
 & IMM-KF & 78.43 & 74.80 & 82.52 & 88.45 & 85.62 & \textbf{1} \\
 & MD-KF  & \textbf{78.66} & \textbf{75.04} & \textbf{82.73}
          & \textbf{88.53} & \textbf{85.89} & \textbf{1} \\
          \hline
 & $\Delta$ MD-KF/Baseline KF
          & \textcolor{blue}{+0.24} & \textcolor{blue}{+0.28} & \textcolor{blue}{+0.20} & \textcolor{blue}{+0.09} & \textcolor{blue}{+0.33} & \textcolor{blue}{$-2$} \\
 & $\Delta$ MD-KF/IMM-KF
          & \textcolor{blue}{+0.23} & \textcolor{blue}{+0.24} & \textcolor{blue}{+0.21} & \textcolor{blue}{+0.08} & \textcolor{blue}{+0.27} & 0 \\
\hline
\multirow{5}{*}{PVCNN~\cite{pv-rcnn_detector}}
 & Baseline KF     & 80.71 & 77.53 & 84.27 & \textbf{89.25} & 87.89 & \textbf{1} \\
 & IMM-KF & 80.78 & 77.69 & 84.27 & 89.17 & 88.17 & \textbf{1} \\
 & MD-KF  & \textbf{81.00} & \textbf{77.93} & \textbf{84.47}
          & 89.20 & \textbf{88.51} & \textbf{1} \\
          \hline
 & $\Delta$ MD-KF/Baseline KF
          & \textcolor{blue}{+0.29} & \textcolor{blue}{+0.40} & \textcolor{blue}{+0.20} & \textcolor{red}{$-0.05$} & \textcolor{blue}{+0.62} & 0 \\
 & $\Delta$ MD-KF/IMM-KF
          & \textcolor{blue}{+0.22} & \textcolor{blue}{+0.24} & \textcolor{blue}{+0.20} & \textcolor{blue}{+0.03} & \textcolor{blue}{+0.34} & 0 \\
\hline
\end{tabular}
\end{table*}

This section compares the performance of the baseline KF, IMM-KF, and the proposed MD-KF on the validation sets of the KITTI and Waymo Open Dataset~\cite{9156973} (WOD). For each dataset, the study compares tracking gain and processing time across methods.

The evaluation on the KITTI dataset includes five different detectors with varying performance to assess performance generalization. As shown in Table~\ref{tab:kitti-five-det-eval}, MD-KF consistently outperforms the baseline KF, with the largest margins of 0.33 in high-order tracking accuracy (HOTA), 0.51 in association accuracy (AssA), and 0.62 in multi-object tracking accuracy (MOTA), while reducing identity switches (IDSw) when a lower detection performance is employed, like PointRCNN (6 → 5) and Second (3 →1). Meanwhile, MD-KF remains competitive with IMM-KF, achieving better performance across most metrics and detectors, with marginal improvements in localization accuracy (LocA); however, IMM-KF further reduces PointRCNN IDSw by 1 compared to MD-KF. Yet, with the VirConv detector, MD-KF improves HOTA and MOTA by 0.46 and 0.35, respectively, while maintaining similar gains across the other detectors. This result shows that MD-KF generally improves state estimation over the baseline KF, while performance remains comparable to IMM-KF. 

Along with the tracking performance evaluation on the KITTI validation set, we record the computational latency of the MD-KF, IMM-KF, and baseline KF across the five detectors. Table~\ref{tab:filter_efficiency_kitti} presents frame-per-second (FPS) throughput in units of $10^3$, along with the speedup over the IMM-KF. The results show that MD-KF provides an average 1.72x runtime speedup over IMM-KF. IMM-KF latency increases as detector performance degrades, as shown for PointRCNN, Second, and PV-CNN, while MD-KF shows higher latency and achieves a 2.15x speedup with PointRCNN. Meanwhile, MD-KF remains closer to the baseline KF throughput than IMM-KF, with its largest speedup over IMM-KF observed for PointRCNN ($2.15\times$) and its smallest for VirConv ($1.50\times$).

\begin{table}[!b]
\centering
\caption{Efficiency performance of the baseline KF, IMM-KF, and the MD-KF on the KITTI validation set across five detectors. The table includes the methods' computational cost, excluding the tracking framework. The latency is measured in $10^3$ FPS. The speedup difference of the MD-KF over the IMM-KF is presented in the last column.}
\label{tab:filter_efficiency_kitti}
\small
\setlength{\tabcolsep}{4pt}
\begin{tabular}{|l|ccc|c|}
\hline
Detector &
KF &
IMM-KF &
MD-KF &
MD/IMM \\
& \multicolumn{3}{c}{FPS ($10^3$)} & Speedup \\
\hline
VirConv   & 131.19 & 68.53 & 102.52 & 1.50$\times$ \\
CasA      & 91.41  & 42.38 & 71.76  & 1.69$\times$ \\
PointRCNN & 50.40  & 20.79 & 44.67  & 2.15$\times$ \\
SECOND    & 88.65  & 39.66 & 68.72  & 1.73$\times$ \\
PV-RCNN   & 90.05  & 37.90 & 72.50  & 1.91$\times$ \\
\hline
Mean      & 90.34  & 41.85 & 72.03  & 1.72$\times$ \\
\hline
\end{tabular}
\end{table}

The tracking results in Table~\ref{tab:kitti-five-det-eval} and the runtime in Table~\ref{tab:filter_efficiency_kitti} suggest that MD-KF improves the performance-computational overhead trade-off relative to IMM-KF, achieving competitive performance while maintaining low latency close to the baseline KF.

\subsubsection{Distance-Range Variation Evaluation}
This study compares the tracking performance of the MD-KF, IMM-KF, and the baseline KF across multiple distance ranges to assess the robustness and limitations of each method. The experiment uses the WOD validation set, and the evaluation results are obtained directly from the official WOD evaluation submission, presented in Table~\ref{tab:wod-eval}. The evaluation includes Level 1 and Level 2 evaluation metrics. Level 1 presents clear objects in the 3D LiDAR point cloud, while Level 2 includes all objects, including challenging objects with limited visibility in the LiDAR point cloud. The evaluation metrics are MOTA, multi-object tracking precision (MOTP; measures localization error), and missed objects (Miss). The evaluation tracks objects in ranges of 0 to 30 meters, 30 to 50 meters, and 50 to +inf meters from the AV. The overall tracking performance is labeled Combined in Table~\ref{tab:wod-eval}.

The evaluation shows an interesting observation about the baseline KF. The baseline KF excels in almost all tracking metrics for close objects (0 to 30 meters) at both levels, compared with IMM-KF and MD-KF, where IMM-KF's tracking performance drops by 0.32 in both MOTA/L1 and MOTA/L2. In contrast, MD-KF drops slightly by 0.08 and 0.07 in MOTA/L1 and MOTA/L2, respectively, showing greater robustness than IMM-KF. As the distance range increases, [30,50) and [50, +inf), MD-KF outperforms the baseline KF and the IMM-KF in all tracking metrics; meanwhile, the IMM-KF and baseline KF compete in performance. 

The combined results across the distance ranges show that MD-KF outperforms the other models across all tracking metrics, while providing $+2475$ FPS higher than IMM-KF.

In this study, particularly beyond 30 meters, MD-KF provides a stable tracking performance gain over the baseline KF and IMM-KF, which begin competing across tracking metrics as distance increases, showing performance instability.

\begin{table*}[!t]
\centering
\caption{\label{tab:wod-eval}Tracking performance evaluation of the baseline KF, IMM-KF, and the proposed MD-KF on the WOD validation set across various range distances. Ranges are presented in meters (m), and processing time is in $10^2$ FPS. The tracking result that combines all ranges is highlighted by “Combined” in the range column. $\Delta$ shows the difference in the combined tracking performance between the MD-KF and the corresponding method. The highest performance for each distance range across the metrics is highlighted in \textbf{bold}. \textcolor{blue}{Blue} highlights MD-KF performance gains, while \textcolor{red}{red} highlights performance degradation.}
\begin{NiceTabular}{|l|l|cccccc|c|}
\hline
\Block{2-1}{Method} & \Block{2-1}{Range (m)}
& \multicolumn{3}{c}{Level 1}
& \multicolumn{3}{c|}{Level 2}
& \Block{2-1}{FPS ($10^2$)} \\
\cmidrule(lr){3-5}\cmidrule(lr){6-8}
& & MOTA $\uparrow$& MOTP $\downarrow$& Miss $\downarrow$& MOTA $\uparrow$& MOTP $\downarrow$& Miss $\downarrow$& \\
\hline
Baseline KF & $[0,30)$       &\textbf{87.14} &\textbf{14.41} &\textbf{8.18} &\textbf{86.41} &\textbf{14.41} &\textbf{8.95} & \\
IMM-KF      & $[0,30)$       &86.82 &14.48 &8.35 &86.09 &14.48 &9.12 & \\
MD-KF       & $[0,30)$       &\underline{87.06} &\underline{14.42} &\underline{8.22} &\underline{86.34} &\underline{14.42} &\underline{8.99} & \\
\hline
Baseline KF & $[30,50)$      &62.59 &\underline{17.34} &\underline{26.38} &59.78 &\underline{17.36} &29.35 & \\
IMM-KF      & $[30,50)$      &\underline{62.63} &17.36 &\underline{26.38} &\underline{59.83} &17.38 &\underline{29.32} & \\
MD-KF       & $[30,50)$      &\textbf{62.80} &\textbf{17.30} &\textbf{26.29} &\textbf{59.99} &\textbf{17.35} &\textbf{28.91} & \\
\hline
Baseline KF & $[50,+\infty)$ &36.33 &\underline{19.32} &48.41 &32.23 &19.39 &\underline{53.27} & \\
IMM-KF      & $[50,+\infty)$ &\underline{36.52} &19.34 &\underline{48.28} &\underline{32.41} &\underline{19.37} &53.59 & \\
MD-KF       & $[50,+\infty)$ &\textbf{36.64} &\textbf{19.28} &\textbf{48.26} &\textbf{32.52} &\textbf{19.35} &\textbf{53.12} & \\
\hline
Baseline KF & Combined       &\underline{62.50} &\underline{16.56} &\underline{27.42} &\underline{58.93} &\underline{16.62} &30.94 & \textbf{159.33}\\
IMM-KF      & Combined       &62.48 &16.61 &27.44 &58.92 &16.66 &\underline{30.93} &119.93 \\
MD-KF       & Combined       &\textbf{62.64} &\textbf{16.55} &\textbf{27.36} &\textbf{59.08} &\textbf{16.60} &\textbf{30.87} & \underline{144.68}\\
\hline
& $\Delta$ MD-KF/Baseline KF & \textcolor{blue}{$+0.14$} & \textcolor{blue}{$-0.01$} & \textcolor{blue}{$-0.06$} & \textcolor{blue}{$+0.15$} & \textcolor{blue}{$-0.02$} & \textcolor{blue}{$-0.07$} & \textcolor{red}{$-14.65$} \\
& $\Delta$ MD-KF/IMM-KF      & \textcolor{blue}{$+0.16$} & \textcolor{blue}{$-0.06$} & \textcolor{blue}{$-0.08$} & \textcolor{blue}{$+0.16$} & \textcolor{blue}{$-0.06$} & \textcolor{blue}{$-0.06$} & \textcolor{blue}{$+24.75$} \\
\hline
\CodeAfter
\begin{tikzpicture}
    % Diagonal across row 3, column 9 (FPS).
    \draw (3-|9) -- (6-|10);
    \draw (6-|9) -- (9-|10);
    \draw (9-|9) -- (12-|10);
    % Uncomment the following line to make an X:
    % \draw (3-|10) -- (4-|9);
\end{tikzpicture}
\end{NiceTabular}
\end{table*}

\subsection{Evalution on Autoware Autonomous Vehicle Stack}
Another evaluation examines the EDGAR~\cite{karle2024edgarautonomousdrivingresearch} autonomous vehicle stack, which relies on the Autoware platform, by integrating RobMOT alongside the MD-KF, IMM-KF, and baseline KF in EDGAR's perception stack using the Robot Operating System (ROS2).  The experiment presents quantitative and qualitative results on the state estimation and tracking performance of the three methods across urban areas and highways in Munich, Germany. The vehicle sensor readings are recorded in a rosbag and used for the study.

\subsubsection{Experiment Protocol: } This study's protocol treats detections from the BEVFusion Autoware detector in the AV as the reference for object locations in the 3D LiDAR point cloud, since the scenario is real-time operation. The study evaluates state estimation from the baseline KF, IMM-KF, and the proposed MD-KF by assessing each method's next-state prediction quality against the detection available at the time. In some experiments, both state prediction and the posterior state (the updated state after the measurement arrives) are also assessed, as shown in Table~\ref{tab:edgar-quantitative_error}. Furthermore, the process time is evaluated for each method. The assessment includes the processing time consumed by the state update step and the state prediction step separately. The tracker settings from the first study (Section~\ref{sec:comp_dataset_benchmarks}) and the tuned parameters in Table~\ref{tab:tuned_params} remain unchanged.

\subsubsection{State Prediction Qualitative Assessment}
\begin{figure*}[!t]
    \centering
    \includegraphics[width=\linewidth]{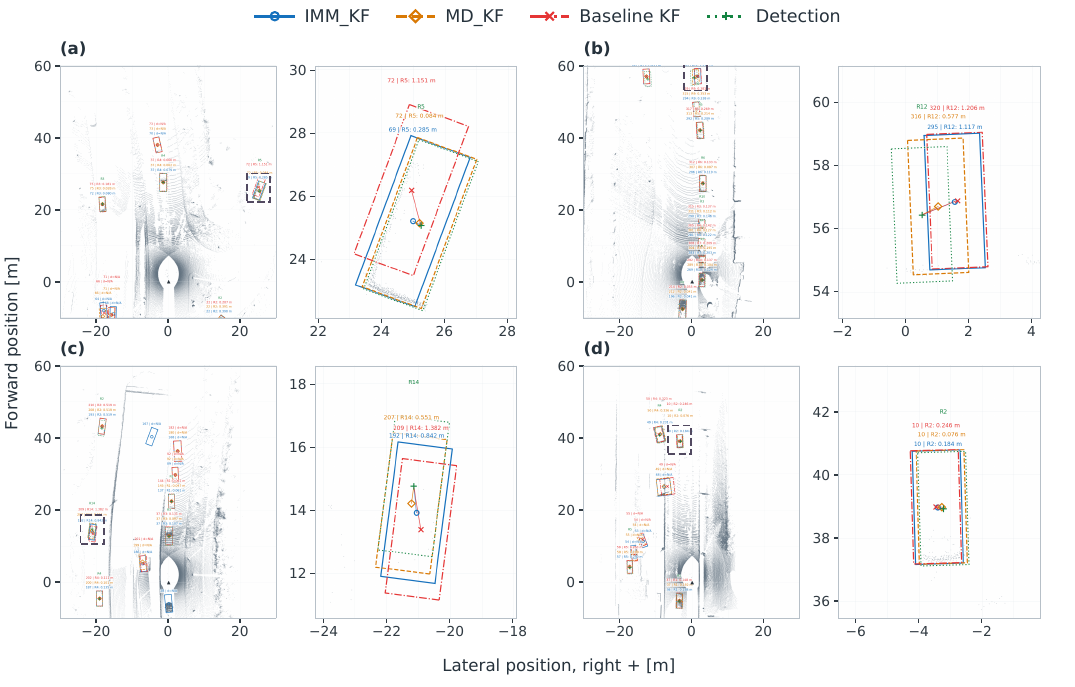}
    \caption{A bird's-eye view of state-estimation performance for the baseline KF, IMM-KF, and proposed MD-KF on the LiDAR 3D point cloud obtained from an EDGAR autonomous vehicle while driving in urban areas in Munich, Germany.  The bounding boxes in this figure come from the methods' state predictions before updating them using the new measurement (posterior). Each lowercase letter shows a separate scenario, with the object of interest zoomed in in an isolated figure on the right. The numerical error above each method’s estimated bounding box is the Euclidean distance between the bounding box center and the center of the detection from EDGAR’s detector. State predictions from MD-KF are highlighted in orange, IMM-KF in blue, and the baseline KF in red. The EDGAR detection is in green.}
    \label{fig:bev-realtime-tracking}
\end{figure*}

\textbf{Scenario (a)} in Figure~\ref{fig:bev-realtime-tracking} shows the AV driving on a highway, with a distant parked car on the right side of the scenario (car: R5). By focusing on car R5, the MD-KF state prediction (orange) aligns best with the car point cloud and has a marginal error of 0.08 meters from the corresponding detection (green). In contrast, the baseline KF prediction (red) drifts about 1.2 meters away from the car's location. The IMM-KF state prediction (blue) remains closer to the car's location than the baseline KF, but still drifts slightly from the true location by about 0.29 meters. 

\textbf{Scenario (b)} in Figure~\ref{fig:bev-realtime-tracking} shows a challenging scenario where a distant, high-speed car (R12) drives on a highway. Focusing on car R12, the zoom-in shows that the detection may drift slightly from the original car point cloud, as shown in Figure~\ref{fig:bev-realtime-tracking} (b) on the right. Nevertheless, the MD-KF predicted state of the R12 car aligns more closely with the car point cloud, with drift of about 0.58 meters from the detection.  In contrast, both the baseline KF and IMM-KF drift more than 1 meter from the car's true location. 

\textbf{Scenario (c)} in Figure~\ref{fig:bev-realtime-tracking} shows a car (R14) on the left side of the AV that was occluded before it reappeared to the LiDAR detector. Upon reappearance, the MD-KF state prediction is closest to the car's true location, with a drift error of 0.55 meters. In contrast, the IMM-KF state prediction drifts by 0.84 meters, and the baseline KF drifts away completely, with an error of 1.38 meters. 

\textbf{Scenario (d)} in Figure~\ref{fig:bev-realtime-tracking} shows a car (R2) driving on a highway, where the detection drifts slightly from its true location based on the point cloud distribution from the backside of the car after zooming in. The MD-KF state prediction aligns best with the car's point cloud, while the baseline KF and IMM-KF drift slightly to the left of the car.

These quantitative experiments highlight only interesting cases observed after examination; however, our observations across many scenarios show competition among the three models in multiple cases, particularly when the object is clearly visible to the detector. Hence, this study presents selected cases where state estimation is considered challenging. A quantitative study is conducted to compare the three methods, as discussed in the next section. 

\begin{table}[!b]
\centering
\caption{Accumulated tracking error for the experiment using 56,997 paired state-estimation samples from the MD-KF, IMM-KF, and baseline KF methods. The error is measured in meters (m) across the metrics. Prediction is the state before updating with the new measurement, while the posterior is the state after updating with it. } 
\label{tab:edgar-quantitative_error}
\renewcommand{\arraystretch}{1.15}
\setlength{\tabcolsep}{5pt}
\begin{tabular}{|l|l|cccc|}
\hline
\textbf{Stage} & \textbf{Method} &
\textbf{Mean} & \textbf{Median} &
\textbf{RMSE} & \textbf{P95} \\
\hline

Prediction
& Baseline
& 0.133 & 0.084 & 0.237 & 0.392 \\
& IMM
& 0.130 & 0.084 & 0.231 & \textbf{0.375} \\

& MD
& \textbf{0.128} & \textbf{0.081} & \textbf{0.230} & 0.379 \\

\hline

Posterior
& Baseline KF
& 0.087 & 0.059 & 0.144 & 0.246 \\
& IMM-KF
& 0.090 & 0.061 & 0.147 & 0.252 \\
& MD-KF
& \textbf{0.084} & \textbf{0.056} & \textbf{0.142} & \textbf{0.240} \\

\hline
\end{tabular}
\end{table}

\subsubsection{State Prediction Quantitative Assessment}
\label{subsub:state-prediction-quantitative}
This experiment evaluates the overall state-estimation performance of the three methods across the full rosbag run. The assessment covers 56,997 samples (detections) and evaluates the three methods' state-estimation performance. To further investigate method performance, the state estimation is evaluated by separating the prediction step from the update step (posterior) when a new measurement arrives.  This separation evaluates the method's ability to provide an accurate prediction and to correct it using measurements. Table~\ref{tab:edgar-quantitative_error} shows the evaluation results across the three methods. The metrics used in the experiment are the alignment error between the state and the detection, summarized across the samples using the mean, median, root mean square error (RMSE), and the 95th percentile.

\begin{figure}[!t]
    \centering
    \includegraphics[width=\linewidth]{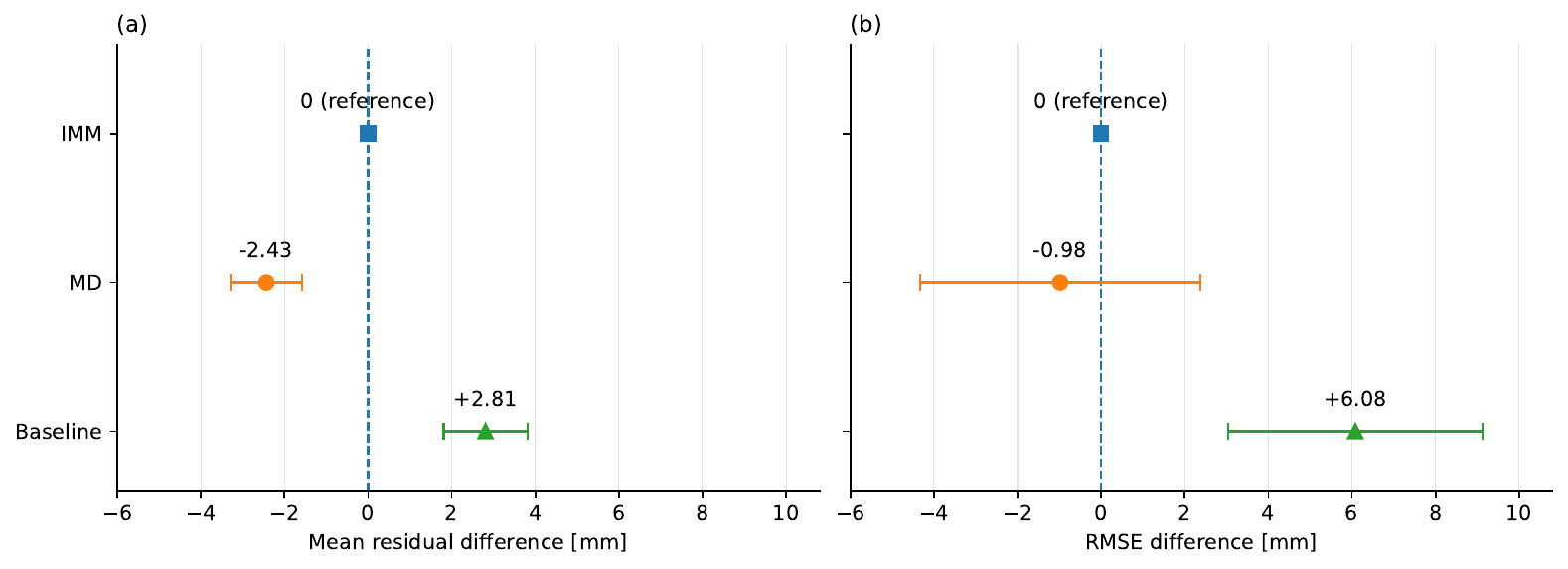}
    \caption{Quantitative comparison of the state prediction alignment error measured over $56{,}997$ paired samples. The plot shows the difference in state estimation error between the baseline KF (green) and the MD-KF (orange), relative to the IMM-KF, which serves as the reference (blue). Positive values indicate higher error than the IMM-KF, while negative values indicate lower error.}
    \label{fig:edgar_qualitative_error}
\end{figure}

Table~\ref{tab:edgar-quantitative_error} shows that, for state prediction, MD-KF achieves lower mean, median, and RMSE alignment errors than IMM-KF. Its 95th-percentile error is slightly higher than IMM-KF's ($0.379$\,m versus $0.375$\,m), indicating comparable tail performance. For the posterior state, MD-KF obtains the lowest error across all four reported statistics. Figure~\ref{fig:edgar_qualitative_error} visualizes the differences in the prediction mean and RMSE relative to IMM-KF: negative values indicate lower aggregate error, whereas positive values indicate higher aggregate error.

\subsubsection{Efficiency Assessment}

\begin{table*}[!b]
\centering
\caption{Tracking robustness under consecutive missing detections: 10-frame and 20-frame gaps on the KITTI validation set using the VirConv detector. $\Delta$ shows the difference in the combined tracking performance between the MD-KF and the corresponding method. The highest performance for frame gap across the metrics is highlighted in \textbf{bold}. \textcolor{blue}{Blue} highlights MD-KF performance gains, while \textcolor{red}{red} highlights performance degradation.}
\label{tab:missing_detection_robustness}
\small
\setlength{\tabcolsep}{4.5pt}
\renewcommand{\arraystretch}{1.08}

\begin{tabular}{|c|l|rrrrrrr|}
\hline
Gap & Method
& HOTA$\uparrow$
& DetA$\uparrow$
& AssA$\uparrow$
& MOTA$\uparrow$
& MOTP$\uparrow$
& IDSW$\downarrow$
& Frag.$\downarrow$ \\
\hline

\multirow{5}{*}{10 frames}
& Baseline KF
& 84.341 & 83.216 & 85.565 & 90.679 & 91.205 & 3 & 18 \\
& IMM-KF
& \textbf{84.788}
& \textbf{83.660}
& \textbf{86.018}
& \textbf{91.312}
& 91.214
& \textbf{2}
& \textbf{11} \\
& MD-KF
& 84.760 & 83.656 & 85.965 & 91.252
& \textbf{91.243}
& 3
& \textbf{11} \\
\hline
& $\Delta$ MD-KF/KF
& \textcolor{blue}{+0.419} & \textcolor{blue}{+0.440} & \textcolor{blue}{+0.400} & \textcolor{blue}{+0.573}
& \textcolor{blue}{+0.038} & 0 & \textcolor{blue}{$-7$} \\
& $\Delta$ MD-KF/IMM-KF
& \textcolor{red}{$-0.028$} & \textcolor{red}{$-0.004$} & \textcolor{red}{$-0.053$} & \textcolor{red}{$-0.060$}
& \textcolor{blue}{+0.029} & \textcolor{red}{+1} & 0 \\
\hline

\multirow{5}{*}{20 frames}
& Baseline KF
& 81.671 & 80.547 & 82.909 & 87.421 & 90.964 & 6 & 21 \\
& IMM-KF
& 83.461 & 81.743 & 85.310 & 88.960 & 90.921 & 3 & 15 \\
& MD-KF
& \textbf{83.820}
& \textbf{82.154}
& \textbf{85.617}
& \textbf{89.486}
& \textbf{90.974}
& \textbf{3}
& \textbf{14} \\
\hline
& $\Delta$ MD-KF/KF
& \textcolor{blue}{+2.149} & \textcolor{blue}{+1.607} & \textcolor{blue}{+2.708} & \textcolor{blue}{+2.065}
& \textcolor{blue}{+0.010} & \textcolor{blue}{$-3$} & \textcolor{blue}{$-7$} \\

& $\Delta$ MD-KF/IMM-KF
& \textcolor{blue}{+0.359} & \textcolor{blue}{+0.411} & \textcolor{blue}{+0.307} & \textcolor{blue}{+0.526}
& \textcolor{blue}{+0.053} & 0 & \textcolor{blue}{$-1$} \\
\hline
\end{tabular}
\end{table*}

\begin{figure}[!t]
    \centering
    \includegraphics[width=\linewidth]{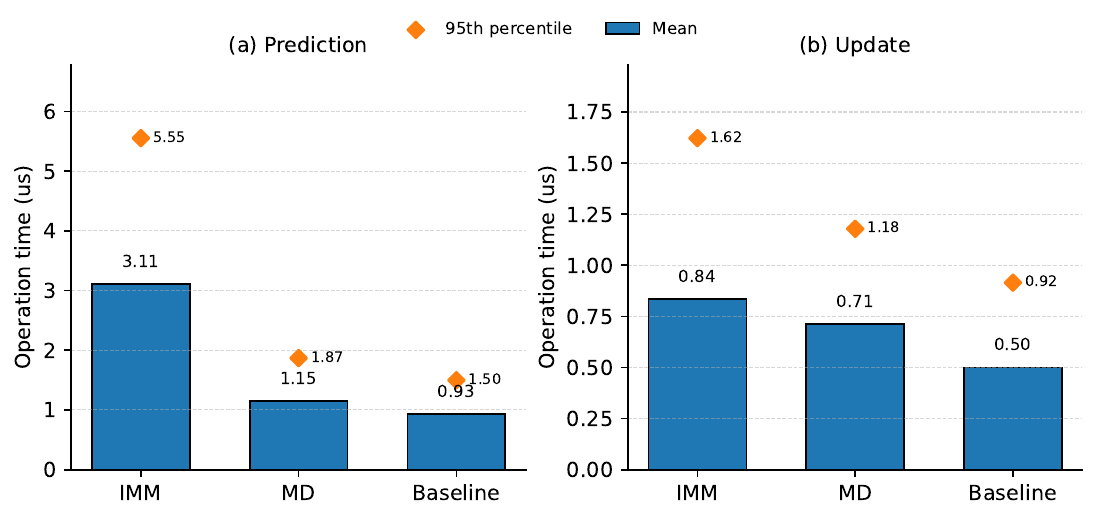}
    \caption{Process time latency for the state update and state prediction steps in the baseline KF, IMM-KF, and proposed MD-KF.}
    \label{fig:edgar_efficiency}
\end{figure}

The efficiency assessment follows the separation strategy between the state prediction and state update steps discussed in Section~\ref{subsub:state-prediction-quantitative}. The operation latency is measured for each stage for the baseline KF, IMM-KF, and the proposed MD-KF across 56.997 samples. Figure~\ref{fig:edgar_efficiency} shows the mean operation time for each stage for the three methods, along with the 95th percentile across samples. The study shows that MD-KF operation time is significantly closer to baseline-KF (0.93→1.15) than IMM-KF, which takes about three times baseline-KF (0.93→3.11) in prediction operation time. Notably, the 95th percentile of IMM-KF operation time spiked in both prediction and update steps, reaching 5.55 in state prediction and 1.62 in state update, respectively. In contrast, MD-KF's 95th percentile is closer to baseline-KF: 1.87 in the prediction step (vs. 1.50 for baseline-KF) and 1.18 in the update step (vs. 0.92 for baseline-KF). This result, combined with MD-KF's performance gain over baseline KF, shows its ability to overcome the performance-efficiency trade-off that baseline KF and IMM-KF struggle with.

\section{Ablation Study}

\subsection{Controlled Missing-Detection Study}
A tracking robustness-under-occlusion study examines the methods' ability to perform precise state estimation for objects with no detections. The study uses the KITTI validation dataset, where systematic detections are removed for selected objects to simulate occlusion. The selection is done for all objects that appear in at least 35 frames, and then their detections are systematically removed for the next 10 or 20 frames. These objects should also appear after the systematic detection removal to evaluate object recovery. This strategy is applied to the entire KITTI validation dataset, and the performance of the baseline KF, IMM-KF, and the proposed MD-KF is evaluated. Table~\ref{tab:missing_detection_robustness} shows results for the two systematic removals: 10 and 20 frames.

In the 10-frame detection removal (Table~\ref{tab:missing_detection_robustness}), MD-KF consistently outperforms the baseline KF, with a 0.419 improvement in HOTA, a 0.573 improvement in MOTA, and reduced trajectory fragments (Frag) (18 → 11). Furthermore, IMM-KF's performance is only marginally better than MD-KF, highlighting the competition between the two models. However, IMM-KF's performance drops significantly compared to MD-KF under prolonged occlusion, as shown in the 20-frame gap in Table~\ref{tab:missing_detection_robustness}.

In the 20-frame detection gap (Table~\ref{tab:missing_detection_robustness}), MD-KF outperforms the baseline KF and the IMM-KF with a noticeable improvement. It improves HOTA by 2.149, MOTA by 2.065, and AssA by 2.708, while reducing the Frag by 7.  Meanwhile, MD-KF improves over IMM-KF by 0.359, 0.526, and 0.411 in HOTA, MOTA, and DetA, respectively. This result suggests that MD-KF not only improves the baseline KF's performance and reaches IMM-KF's performance but also outperforms IMM-KF during prolonged occlusions. 

\subsection{Adaptive Motion Weights Contribution Study}
\begin{table*}[!t]
\centering
\caption{Ablation of the adaptive motion-dynamics weights on the KITTI validation set using VirConv. A weight marked as disabled is fixed to one, removing its adaptive effect while retaining the underlying kinematic state. The final row reports the improvement of the complete MD-KF over the baseline-KF with $\mathbf{W}_t=\mathbf{I}$. $\Delta$ shows the difference in tracking performance between the fully enabled MD-KF weights and the disabled MD-KF weights (the baseline KF). The best performance across metrics is highlighted in \textbf{bold}. \textcolor{blue}{Blue} highlights MD-KF with fully enabled weights performance gains, while \textcolor{red}{red} highlights performance degradation.}
\label{tab:adaptive_weight_ablation}
\small
\setlength{\tabcolsep}{5pt}
\renewcommand{\arraystretch}{1.08}

\begin{tabular}{|l|ccccccc|}
\hline
Adaptive weights enabled
& HOTA$\uparrow$
& DetA$\uparrow$
& AssA$\uparrow$
& MOTA$\uparrow$
& MOTP$\uparrow$
& IDSW$\downarrow$
& Frag.$\downarrow$ \\
\hline
None ($\mathbf{W}_t=\mathbf{I}$)
& 85.965 & 83.988 & 88.075 & 91.121 & 91.609 & 1 & 8 \\

$w^j$ only
& 86.043 & 84.097 & 88.123 & 91.252 & 91.618 & 1 & 7 \\

$w^a$ and $w^j$
& 86.143 & 84.201 & 88.218 & 91.347 & 91.633
& \textbf{1} & \textbf{6} \\

Full: $w^v$, $w^a$, and $w^j$
& \textbf{86.296}
& \textbf{84.369}
& \textbf{88.356}
& \textbf{91.526}
& \textbf{91.635}
& \textbf{1}
& \textbf{6} \\
\hline
$\Delta$ Full/$\mathbf{W}_t=\mathbf{I}$
& \textcolor{blue}{+0.331} & \textcolor{blue}{+0.381} & \textcolor{blue}{+0.281} & \textcolor{blue}{+0.405} & \textcolor{blue}{+0.026} & 0 & \textcolor{blue}{$-2$} \\

\hline
\end{tabular}
\end{table*}
\begin{figure}[!b]
    \centering
    \includegraphics[width=\linewidth]{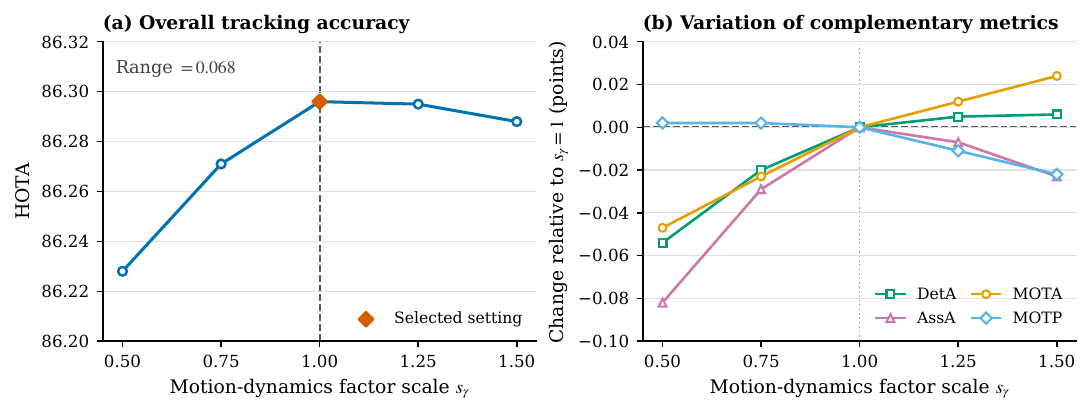}
    \caption{ A sensitivity of the MD-KF normalization factors scaled by $s_\gamma$ on the KITTI validation set. (a) shows how tracking performance changes with scaling on the HOTA metric. (b) shows how DetA, MOTA, AssA, and MOTP vary across the scaling. The scaling $s_\gamma=1$ is the original tuned factors without scaling. }
    \label{fig:sensitive_norm_factors}
\end{figure}
Because MD-KF quantifies changes in position, velocity, and acceleration to determine the contribution of the motion model terms (velocity, acceleration, and jerk) and then corrects the state prediction accordingly, this study quantifies each adaptive motion weight's contribution in the MD-KF. The experiment deactivates all weights by setting the weight matrix $ W_t$ to the identity matrix ($I$). Then, the adaptive weights are gradually activated, starting with the jerk weight $W^j$. The study uses the KITTI validation dataset, and the results are presented in Table~\ref{tab:adaptive_weight_ablation}.

As shown in Table~\ref{tab:adaptive_weight_ablation}, including $w^j$ in the baseline KF improves tracking performance across all metrics, particularly by reducing Frag (8 → 7) and producing a more consistent state-estimated trajectory.  By activating $ w^a$, tracking performance improves further across all metrics, with Frag further reduced (7→6).  Including the last weight $w^v$ also improves all metrics, suggesting incremental gains from integrating motion-dynamics weights into the baseline KF. 

These results show the impact of the proposed MD-KF methodology on the baseline KF, supporting the importance of adaptive weights to correct drift in state estimation caused by variations in the object’s motion, thereby improving overall tracking performance.

\subsection{Parameter Sensitivity Study}
\begin{figure}[!b]
    \centering
    \includegraphics[width=\linewidth]{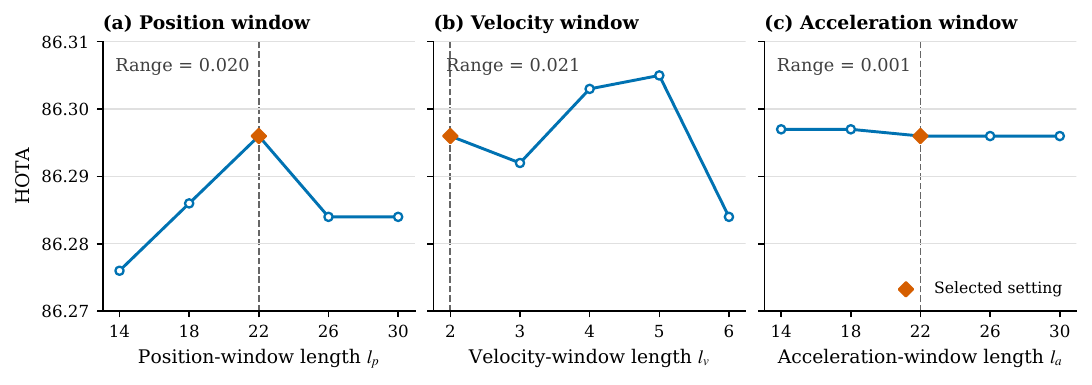}
    \caption{Sensitivity of the MD-KF window length to the position $l_p$ (a), velocity $l_v$ (b), and acceleration $l_a$ (c). Tracking performance variation is measured using the HOTA metric on the KITTI validation set. The selected settings are highlighted in an orange diamond in each plot.}
    \label{fig:sensitive_window}
\end{figure}

MD-KF introduces window-length and normalization-factor parameters to capture and adapt motion evidence from objects. These parameters are tuned on the KITTI training dataset using the CasA detector, and the selected values are listed in Table~\ref{tab:tuned_params}, Section~\ref{sub:exp_protocol}. To examine how sensitive these parameters are and ensure there is no hard tuning to improve tracking performance, a sensitivity study is conducted across the window-length and normalization-factor parameters. The study is done using the VirConv detector on the KITTI validation dataset. 

Initially, a sensitivity study is conducted on the normalization factors ($\gamma^p, \gamma^v, \gamma^a$) by applying a scale factor \[s_\gamma\in\{0.50,\;0.75,\;1.00,\;1.25,\;1.50\}.\] to the factors and evaluating the change in the tracking performance metrics. This study uses HOTA, MOTA, MOTP, AssA, and DetA as tracking metrics. The study results are plotted in Figure~\ref{fig:sensitive_norm_factors}.

The study shows that scaling the factors from 0.5 to 1.50 results in a negligible change in HOTA, $0.068$, while the variations in the remaining metrics remain within $[-0.08,\,0.03]$ points. More importantly, the line plots show HOTA peaks on the original tuned factors when the scale factor $s_\gamma=1$, while the other tracking metrics are balanced at $s_\gamma=1$, where AssA and MOTP metrics begin to degrade as the factor increases.

The second study evaluates MD-KF's sensitivity to the temporal-window lengths $(l_p,l_v,l_a)$ by varying one window at a time while holding the remaining two at their selected values. Specifically, we vary the position and acceleration window lengths over $\{14,18,22,26,30\}$ and the velocity window length over $\{2,3,4,5,6\}$. The experiment evaluates the performance on the HOTA metric. The study outcomes are presented in Figure~\ref{fig:sensitive_window}.

As shown in Figure~\ref{fig:sensitive_window}, HOTA changes very little as the window length varies, not exceeding 0.021. The velocity window shows the largest change (0.021), while the acceleration window shows the smallest change (0.001). Together, the results indicate that MD-KF operates within a broad stable parameter region and does not require narrowly optimized parameter values to maintain tracking performance.
\section{Conclusion}
This paper addresses the limitation of the conventional Kalman filter (Baseline KF) in accounting for tracked-object motion variation and overcomes the performance-efficiency tradeoff between the interacting multiple-model Kalman filter (IMM-KF) and the Baseline KF by proposing the motion dynamics Kalman filter (MD-KF). MD-KF captures changes in object motion as motion evidence and adaptively adjusts the contribution of the motion-model components, enabling adaptation to object motion. MD-KF consistently outperformed the baseline KF across multiple detectors and benchmark datasets, while matching IMM-KF performance. The experiments show that the baseline KF performs well with near objects, whereas IMM-KF’s performance degrades; MD-KF mitigates this degradation and matches the baseline KF's performance with near objects. Meanwhile, MD-KF outperforms the others in tracking highly distant objects. A controlled occlusion study also shows substantial improvement in tracking performance under prolonged occlusion, whereas the baseline KF's performance drops significantly, while IMM-KF achieves moderate performance. Further experiments on an autonomous vehicle Autoware stack show comparable tracking performance between MD-KF and IMM-KF, while MD-KF achieves the overall lowest state estimation error. MD-KF maintains the baseline KF's efficiency, achieving 1.72 lower average latency than IMM-KF.  A sensitivity study on MD-KF shows that it operates over a broad range of stable parameters while maintaining tracking performance, which simplifies tuning. MD-KF's performance gain over the baseline KF, along with its efficiency over IMM-KF, makes it a strong state-estimation approach for online multi-object tracking in autonomous systems.

\section*{Acknowledgments}
The authors thank the Autonomous Vehicle Lab~\footnote{https://www.mos.ed.tum.de/en/ftm/labs/autonomous-vehicle/} and the AV perception team at the Chair of Automotive Technology, Technical University of Munich, for supporting the autonomous-vehicle experiments and recording the vehicle data used in this study.

\bibliographystyle{IEEEtran}
\bibliography{ref}

@INPROCEEDINGS{intro_tra_by_det_kim_eager,
  author={Kim, Aleksandr and Ošep, Aljoša and Leal-Taixé, Laura},
  booktitle={2021 IEEE International Conference on Robotics and Automation (ICRA)}, 
  title={EagerMOT: 3D Multi-Object Tracking via Sensor Fusion}, 
  year={2021},
  volume={},
  number={},
  pages={11315-11321},
  doi={10.1109/ICRA48506.2021.9562072}}

@INPROCEEDINGS{nagy,
  author={Nagy, Mohamed and Khonji, Majid and Dias, Jorge and Javed, Sajid},
  booktitle={2023 IEEE International Conference on Robotics and Automation (ICRA)}, 
  title={DFR-FastMOT: Detection Failure Resistant Tracker for Fast Multi-Object Tracking Based on Sensor Fusion}, 
  year={2023},
  volume={},
  number={},
  pages={827-833},
  doi={10.1109/ICRA48891.2023.10160328}}

@INPROCEEDINGS{Geiger2012CVPR,
  author = {Andreas Geiger and Philip Lenz and Raquel Urtasun},
  title = {Are we ready for Autonomous Driving? The KITTI Vision Benchmark Suite},
  booktitle = {Conference on Computer Vision and Pattern Recognition (CVPR)},
  year = {2012}
}

@INPROCEEDINGS{VirConvDet_CVPR_23,
  author={Wu, Hai and Wen, Chenglu and Shi, Shaoshuai and Li, Xin and Wang, Cheng},
  booktitle={2023 IEEE/CVF Conference on Computer Vision and Pattern Recognition (CVPR)}, 
  title={Virtual Sparse Convolution for Multimodal 3D Object Detection}, 
  year={2023},
  volume={},
  number={},
  pages={21653-21662},
  doi={10.1109/CVPR52729.2023.02074}}

@ARTICLE{casA_detector,
  author={Wu, Hai and Deng, Jinhao and Wen, Chenglu and Li, Xin and Wang, Cheng and Li, Jonathan},
  journal={IEEE Transactions on Geoscience and Remote Sensing}, 
  title={CasA: A Cascade Attention Network for 3-D Object Detection From LiDAR Point Clouds}, 
  year={2022},
  volume={60},
  number={},
  pages={1-11},
  doi={10.1109/TGRS.2022.3203163}}

@InProceedings{point_rcnn_detector,
    author = {Shi, Shaoshuai and Wang, Xiaogang and Li, Hongsheng},
    title = {PointRCNN: 3D Object Proposal Generation and Detection From Point Cloud},
    booktitle = {The IEEE Conference on Computer Vision and Pattern Recognition (CVPR)},
    month = {June},
    year = {2019}
}

@InProceedings{pv-rcnn_detector,
author = {Shi, Shaoshuai and Guo, Chaoxu and Jiang, Li and Wang, Zhe and Shi, Jianping and Wang, Xiaogang and Li, Hongsheng},
title = {PV-RCNN: Point-Voxel Feature Set Abstraction for 3D Object Detection},
booktitle = {Proceedings of the IEEE/CVF Conference on Computer Vision and Pattern Recognition (CVPR)},
month = {June},
year = {2020}
}

@article{second_detector,
  title={Second: Sparsely embedded convolutional detection},
  author={Yan, Yan and Mao, Yuxing and Li, Bo},
  journal={Sensors},
  volume={18},
  number={10},
  pages={3337},
  year={2018},
  publisher={MDPI}
}

@INPROCEEDINGS{Rethink_mot,
  author={Wang, Leichen and Zhang, Jiadi and Cai, Pei and Lil, Xinrun},
  booktitle={2023 IEEE International Conference on Robotics and Automation (ICRA)}, 
  title={Towards Robust Reference System for Autonomous Driving: Rethinking 3D MOT}, 
  year={2023},
  volume={},
  number={},
  pages={8319-8325},
  doi={10.1109/ICRA48891.2023.10160645}}

@ARTICLE{CasTrack,
  author={Wu, Hai and Han, Wenkai and Wen, Chenglu and Li, Xin and Wang, Cheng},
  journal={IEEE Transactions on Intelligent Transportation Systems}, 
  title={3D Multi-Object Tracking in Point Clouds Based on Prediction Confidence-Guided Data Association}, 
  year={2022},
  volume={23},
  number={6},
  pages={5668-5677},
  doi={10.1109/TITS.2021.3055616}}

@INPROCEEDINGS{9156973,
  author={Sun, Pei and Kretzschmar, Henrik and Dotiwalla, Xerxes and Chouard, Aurélien and Patnaik, Vijaysai and Tsui, Paul and Guo, James and Zhou, Yin and Chai, Yuning and Caine, Benjamin and Vasudevan, Vijay and Han, Wei and Ngiam, Jiquan and Zhao, Hang and Timofeev, Aleksei and Ettinger, Scott and Krivokon, Maxim and Gao, Amy and Joshi, Aditya and Zhang, Yu and Shlens, Jonathon and Chen, Zhifeng and Anguelov, Dragomir},
  booktitle={2020 IEEE/CVF Conference on Computer Vision and Pattern Recognition (CVPR)}, 
  title={Scalability in Perception for Autonomous Driving: Waymo Open Dataset}, 
  year={2020},
  volume={},
  number={},
  pages={2443-2451},
  doi={10.1109/CVPR42600.2020.00252}}

@inproceedings{Pang2021SimpleTrackUA,
  title={SimpleTrack: Understanding and Rethinking 3D Multi-object Tracking},
  author={Ziqi Pang and Zhichao Li and Naiyan Wang},
  booktitle={ECCV Workshops},
  year={2021},
  url={https://api.semanticscholar.org/CorpusID:244346157}
}

@article{Weng20193DMT,
  title={3D Multi-Object Tracking: A Baseline and New Evaluation Metrics},
  author={Xinshuo Weng and Jianren Wang and David Held and Kris Kitani},
  journal={2020 IEEE/RSJ International Conference on Intelligent Robots and Systems (IROS)},
  year={2019},
  pages={10359-10366},
  url={https://api.semanticscholar.org/CorpusID:220686905}
}

@INPROCEEDINGS{9626850,
  author={Reich, Andreas and Wuensche, Hans-Joachim},
  booktitle={2021 IEEE 24th International Conference on Information Fusion (FUSION)}, 
  title={Monocular 3D Multi-Object Tracking with an EKF Approach for Long-Term Stable Tracks}, 
  year={2021},
  volume={},
  number={},
  pages={1-7},
  doi={10.23919/FUSION49465.2021.9626850}}

@article{wang2021immortal,
  title={Immortal tracker: Tracklet never dies},
  author={Wang, Qitai and Chen, Yuntao and Pang, Ziqi and Wang, Naiyan and Zhang, Zhaoxiang},
  journal={arXiv preprint arXiv:2111.13672},
  year={2021}
}

@article{zhu2022msa,
  title={MSA-MOT: Multi-Stage Association for 3D Multimodality Multi-Object Tracking},
  author={Zhu, Ziming and Nie, Jiahao and Wu, Han and He, Zhiwei and Gao, Mingyu},
  journal={Sensors},
  volume={22},
  number={22},
  pages={8650},
  year={2022},
  publisher={MDPI}
}

@article{he20243d,
  title={3D multi-object tracking based on informatic divergence-guided data association},
  author={He, Jiawei and Fu, Chunyun and Wang, Xiyang and Wang, Jianwen},
  journal={Signal Processing},
  volume={222},
  pages={109544},
  year={2024},
  publisher={Elsevier}
}

@inproceedings{scheidegger2018mono,
  title={Mono-camera 3d multi-object tracking using deep learning detections and pmbm filtering},
  author={Scheidegger, Samuel and Benjaminsson, Joachim and Rosenberg, Emil and Krishnan, Amrit and Granstr{\"o}m, Karl},
  booktitle={2018 IEEE Intelligent Vehicles Symposium (IV)},
  pages={433--440},
  year={2018},
  organization={IEEE}
}

@article{na2022adaptive,
  title={Adaptive target tracking with interacting heterogeneous motion models},
  author={Na, Ki-In and Choi, Sunglok and Kim, Jong-Hwan},
  journal={IEEE Transactions on Intelligent Transportation Systems},
  volume={23},
  number={11},
  pages={21301--21313},
  year={2022},
  publisher={IEEE}
}

@article{yan2013maneuvering,
  title={A Maneuvering Target Tracking Algorithm Based on the Interacting Multiple Models},
  author={Yan-Chang, Liu and Xian-Gang, Zuo},
  journal={TELKOMNIKA Indonesian Journal of Electrical Engineering},
  volume={11},
  number={7},
  pages={3997--4003},
  year={2013}
}

@inproceedings{sani2024sensor,
  title={Sensor-Agnostic Graph-Aware Kalman Filter for Multi-Modal Multi-Object Tracking},
  author={Sani, Depanshu and Iyer, Anirudh and Rai, Prakhar and Anand, Saket and Srivastava, Anuj and Kalyanaraman, Kaushik},
  booktitle={International Conference on Pattern Recognition},
  pages={380--398},
  year={2024},
  organization={Springer}
}

@article{10.1115/1.3662552,
    author = {Kalman, R. E.},
    title = {A New Approach to Linear Filtering and Prediction Problems},
    journal = {Journal of Basic Engineering},
    volume = {82},
    number = {1},
    pages = {35-45},
    year = {1960},
    month = {03},
    issn = {0021-9223},
    doi = {10.1115/1.3662552},
    url = {https://doi.org/10.1115/1.3662552},
    eprint = {https://asmedigitalcollection.asme.org/fluidsengineering/article-pdf/82/1/35/5518977/35_1.pdf},
}

@ARTICLE{11249796,
  author={Liu, Hengyuan and Chen, Zhong and Du, Zhenzhen and Li, Hui and Ai, Xiaoxue},
  journal={IEEE Transactions on Intelligent Transportation Systems}, 
  title={3D Multi-Object Tracking Driven by Multi-Level Association and Intelligent Filtering}, 
  year={2026},
  volume={27},
  number={1},
  pages={1442-1457},
  doi={10.1109/TITS.2025.3625426}}

@INPROCEEDINGS{11246301,
  author={Liu, Xiaohong and Zhao, Xulong and Liu, Gang and Wu, Zili and Wang, Tao and Meng, Lei and Wang, Yuhan},
  booktitle={2025 IEEE/RSJ International Conference on Intelligent Robots and Systems (IROS)}, 
  title={IMM-MOT: A Novel 3D Multi-object Tracking Framework with Interacting Multiple Model Filter}, 
  year={2025},
  volume={},
  number={},
  pages={20540-20547},
  doi={10.1109/IROS60139.2025.11246301}}

@misc{karle2024edgarautonomousdrivingresearch,
      title={EDGAR: An Autonomous Driving Research Platform -- From Feature Development to Real-World Application}, 
      author={Phillip Karle and Tobias Betz and Marcin Bosk and Felix Fent and Nils Gehrke and Maximilian Geisslinger and Luis Gressenbuch and Philipp Hafemann and Sebastian Huber and Maximilian Hübner and Sebastian Huch and Gemb Kaljavesi and Tobias Kerbl and Dominik Kulmer and Tobias Mascetta and Sebastian Maierhofer and Florian Pfab and Filip Rezabek and Esteban Rivera and Simon Sagmeister and Leander Seidlitz and Florian Sauerbeck and Ilir Tahiraj and Rainer Trauth and Nico Uhlemann and Gerald Würsching and Baha Zarrouki and Matthias Althoff and Johannes Betz and Klaus Bengler and Georg Carle and Frank Diermeyer and Jörg Ott and Markus Lienkamp},
      year={2024},
      eprint={2309.15492},
      archivePrefix={arXiv},
      primaryClass={cs.RO},
      url={https://arxiv.org/abs/2309.15492}, 
}

@ARTICLE{nagy2024robmotrobust3dmultiobject,
  author={Nagy, Mohamed and Werghi, Naoufel and Hassan, Bilal and Dias, Jorge and Khonji, Majid},
  journal={IEEE Transactions on Intelligent Transportation Systems}, 
  title={RobMOT: 3D Multi-Object Tracking Enhancement Through Observational Noise and State Estimation Drift Mitigation in LiDAR Point Clouds}, 
  year={2025},
  volume={26},
  number={10},
  pages={16047-16059},
  doi={10.1109/TITS.2025.3581980}}

@INPROCEEDINGS{imm_kf_mot24,
  author={Li, Qilin and Zhang, Zhenhai and He, Guang and Hu, Xuehai and Kang, Xiao},
  booktitle={2024 IEEE International Conference on Unmanned Systems (ICUS)}, 
  title={3D Multi-Object Tracking for Autonomous Driving Based on IMM - EKF and Re- Identification}, 
  year={2024},
  volume={},
  number={},
  pages={1197-1202},
  doi={10.1109/ICUS61736.2024.10839925}}


\end{document}